# An open benchmark for machine learning-based polymer property prediction

Robert W. Learsch[1], Nicholas Liesen[1], Daniel S. Levine[2], Anna M. Hiszpanski[1,*], Evan R. Antoniuk[1,*]

*Corresponding authors (hiszpanski2@llnl.gov, antoniuk1@llnl.gov)

[1]Lawrence Livermore National Laboratory, [2]FAIR at Meta

## Abstract

Polymer property prediction lacks open, standardized benchmarks that enable rigorous comparison of machine-learning methods, with existing resources covering only a narrow fraction of polymer architectures, such as homopolymers. We introduce Polymer Benchmark 2026 (PolyBench26), an open dataset comprising nearly 250,000 polymer-property datapoints across eight physical properties, including data from experimental measurements, density functional theory, and molecular dynamics. The benchmark supports four evaluation tasks across homopolymers and alternating, random, and block copolymers: in-distribution property prediction, dataset-size scaling, repeat-unit complexity, and transfer to held-out polymer architectures. We compare language model, graph-based, and descriptor-based approaches and find graph-based models provide the lowest errors in property prediction, retain their advantage across the evaluated training-set sizes, and remain robust to increasing repeat-unit complexity. PolyBench26 provides a reproducible foundation for developing models for the increasingly complex polymer design space. The PolyBench26 benchmark is available open-source at https://github.com/rlearsch/PolymerBenchmark2026.

## Introduction

Unlike other areas of chemistry where standardized benchmarks such as MatBench and MoleculeNet enable rigorous tracking of machine learning (ML) model performance, polymer informatics lacks open, standardized datasets and benchmarks for property prediction.[1,2] This absence obscures the state of the art models and makes it difficult to determine which modeling advances represent genuine progress.[3,4] Several recent studies have identified the lack of comprehensive polymer databases as a major barrier to ML-driven polymer design.[5–9] However, prior efforts have focused almost exclusively on homopolymers, which contain a single repeat unit.[10–12]

Homopolymers represent only a fraction of polymer design space. Many technologically important materials are copolymers, which contain two or more repeat units arranged in specific architectures.[13–16] Their organization into random, block, or alternating structures can substantially influence thermal, mechanical, and electronic properties.[13,14,17–20] To enable the discovery of higher-performing polymers, ML models must accurately represent and predict properties across increasingly complex polymer architectures, including copolymers, graft polymers, and polymer networks.[18,21,22]

The community has begun addressing these challenges through open datasets[23–26] such as VIPEA,[18] polyVERSE,[27–29] OpenPoly,[30] and PolyMetriX.[31] However, these resources remain limited in their coverage of polymer architectures, properties, and representations. VIPEA consists of about 90,000 DFT-calculated IP and EA data points and strictly contains copolymers. polyVERSE, which is under active development, now provides approximately 30,000 experimental, DFT-calculated, and simulation property values spanning 59 quantities.[28,29] For PolyBench26, we included the approximately 700 DFT-calculated homopolymer electron-affinity (EA) and ionization potential (IP) values from polyVERSE to provide electronic-property benchmarks that complement the copolymer data in VIPEA. OpenPoly spans 26 homopolymer properties, but contains only 30 to 443 examples for any individual property.[30] PolyMetriX provides open-access experimental data for more than 7,000 homopolymer glass transition temperatures and compares polyBERT with Morgan fingerprints, but does not evaluate graph-based or text-based models.[31] Thus, the field still lacks a unified benchmark that combines homopolymers and copolymers, spans experimental and computational properties, and supports systematic comparison of modern polymer representations and model architectures.

In this work, we address this gap by developing the Polymer Benchmark 2026 (PolyBench26), comprising approximately 250,000 polymer-property datapoints across eight physical properties and multiple polymer architectures, including homopolymers, alternating copolymers, block copolymers, and random copolymers. We evaluate three broad classes of polymer vectorization schemes and four benchmark tasks representing common polymer-property prediction scenarios: in-distribution prediction, dataset-size scaling, prediction across increasing repeat-unit complexity, and generalization to copolymer architectures absent from the training data. To explore how the choice of model architecture impacts the polymer property prediction accuracy, we compare language model,[32] graph-based, and descriptor-based approaches. Graph-based models perform best for homopolymers and alternating copolymers, achieving superior in-distribution performance across dataset sizes and greater robustness to increasing repeat-unit complexity. Altogether, PolyBench26 provides an open-source framework for evaluating polymer ML models, allowing the polymer informatics community to identify representations that support reliable prediction across increasingly complex polymer design spaces.

# Results

## Dataset composition and models evaluated

In this work, we present PolyBench26, a large collection of open polymer properties datasets and benchmark tasks, summarized in Figure 1. In Task 1, we assess in-distribution property prediction performance by testing text-based, graph-based, and descriptor-based models on polymers compositionally and structurally similar to those represented in the training data. Because the text-based models showed substantially lower predictive performance, Tasks 2-4 focus on graph-based and descriptor-based approaches. Task 2 evaluates performance as a function of the total number of training datapoints, providing insights into model performance in the low-data regime. Task 3 examines how models handle repeat units of increasing structural complexity by evaluating prediction error as a function of chemically distinct substituents in the repeat unit. Finally, in Task 4 we assess how models can transfer knowledge across polymer architecture by performing architecture-held-out evaluations in which random and block copolymers were excluded from training.

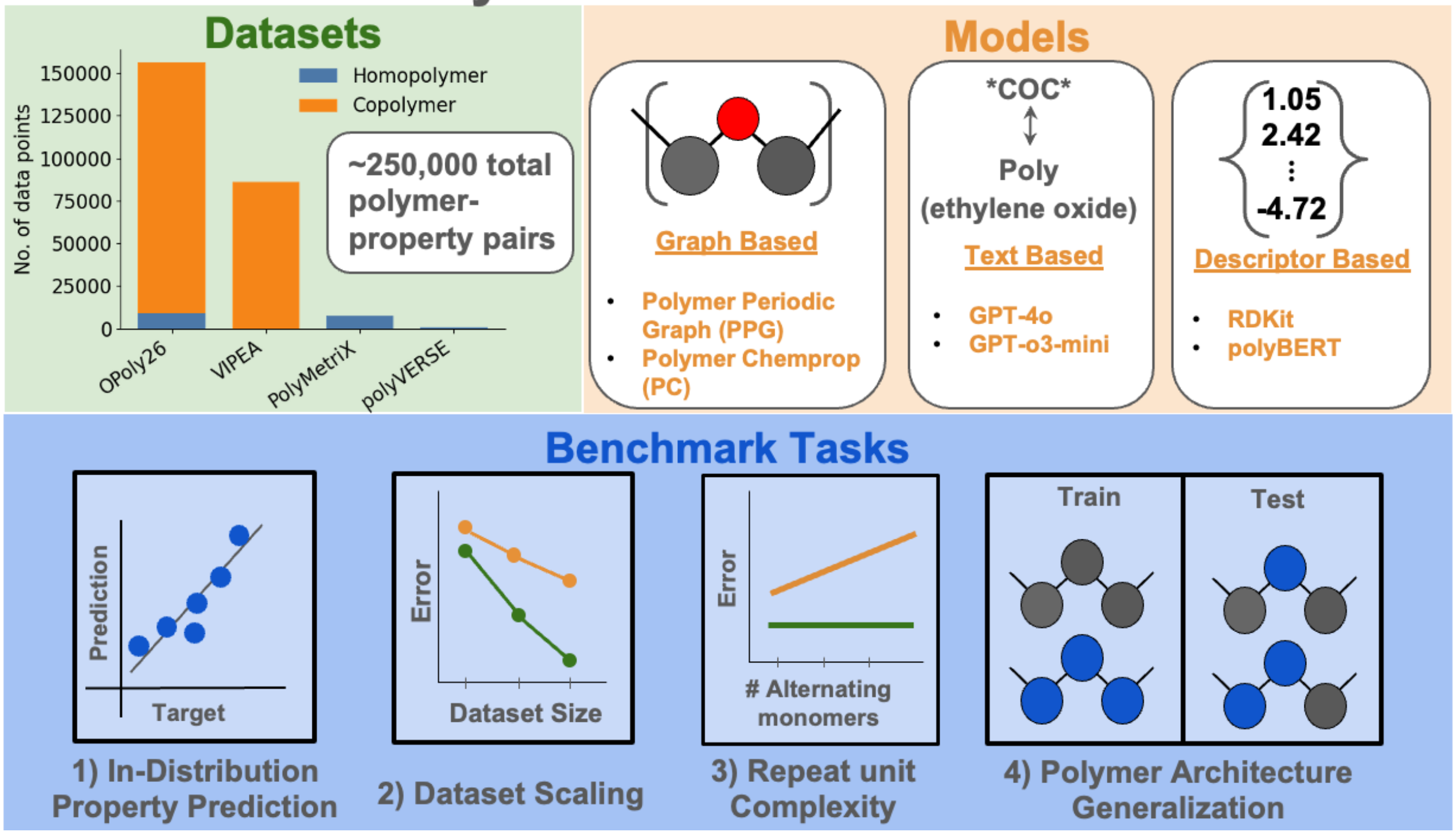

**Figure 1.** Overview of PolyBench26. The datasets of PolyBench26 are collected from existing open-source polymer datasets, as well as newly extracted data from OPoly26, and spans both copolymer and homopolymer architectures. We systematically evaluate the polymer prediction performance of graph-based, text-based, and descriptor-based approaches. Models are evaluated against four different benchmark tasks, designed to probe different aspects of model performance.

As summarized in Table 1, PolyBench26 combines existing polymer-property datasets, including VIPEA,[18] PolyMetriX,[31] and polyVERSE.[27–29] Recently, the OPoly26 dataset of 94,000 unique amorphous polymers was released to enable the training of machine-learned interatomic potentials (MLIPs), but without any additional property data. In this work, we utilize the OPoly26 molecular dynamics (MD) trajectories to extract an additional 155,930 polymer property datapoints. Specifically, we utilize the methods implemented in the RadonPy package to calculate density, specific heat ($C_p$), refractive index, and radius-of-gyration from the existing OPoly26 MD trajectories (see Methods).[22] We compare these calculated values against experimental properties and find good agreement for density (Figure S4) and refractive index (Figure S3), validating their use in benchmarking ML models. We note that the radius of gyration calculation depends on the finite chain length used in the MD simulations and therefore is not intended as a robust physical observable; rather, we include it in the benchmark as a test of each model's ability to learn and predict polymer chain statistics based on the inputs of a MD simulation. The default methodology of RadonPy for calculating heat capacity was found to give poor agreement when compared against experimental heat capacities (Figure S5). To give significantly more accurate heat capacity values, we perform additional MD calculations that specifically account for frozen oscillation modes (see Methods). We find that accounting for frozen oscillation modes[33] reduces the RMSE between MD-calculated $C_v$ and experimentally measured $C_p$ from $1.9$ to $0.4\ \mathrm{Jg^{-1}K^{-1}}$, a reduction of approximately 79% (Figures S5 and S6).[34] We include both the classical calculation of $C_p$ for compatibility with RadonPy alongside the quantum-corrected $C_v$ calculations.

**Table 1**. Composition of the PolyBench26 dataset**.** Summary of the polymer properties, measurement sources, polymer architectures, number of datapoints, and corresponding data sources included in the benchmark. “Alt.” denotes alternating copolymers.

| Property (units) | Measurement Source | Polymer Architecture | Data points | Data Source |
|---|---|---|---|---|
| Glass transition temperature ($T_g$) (°C) | Experiment | Homopolymers | 7,362 | PolyMetriX[31] |
| Electron affinity (EA) (eV) | DFT | Homopolymers | 368 | polyVERSE[27–29] |
| | | Copolymers (Alt., Block, Random) | 42,966 | VIPEA[18] |
| Ionization potential (IP) (eV) | DFT | Homopolymers | 370 | polyVERSE |
| | | Copolymers (Alt., Block, Random) | 42,966 | VIPEA |
| Refractive index (n) | DFT | Homopolymers | 1,715 | OPoly26[16] |
| | | Copolymers (Alt., Random) | 35,628 | OPoly26 |
| Density (g/mL) | MD | Homopolymers | 1,800 | OPoly26 |
| | | Copolymers (Alt., Random) | 37,133 | OPoly26 |
| Specific heat ($C_p$) (J/kg K) | MD | Homopolymers | 1,800 | OPoly26 |
| | | Copolymers (Alt., Random) | 37,133 | OPoly26 |
| Specific heat ($C_v$) (J/g K) | MD | Homopolymers | 1,788 | OPoly26 |
| Radius of gyration (nm) | MD | Homopolymers | 1,800 | OPoly26 |

| | | Copolymers (Alt., Random) | 37,133 | OPoly26 |
|---|---|---|---|---|

We focus on three broad classes of vectorization schemes used in polymer-property prediction, explained in Table 2. We evaluate text-based language models GPT-4o and GPT-o3-mini; graph-based models Polymer Chemprop (PC) and Polymer Periodic Graph (PPG); and descriptor-based models polyBERT and a Random Forest (RF) with RDKit descriptors, used as a baseline. Every model except Polymer Chemprop uses a polymer SMILES (PSMILES) representation of the polymer. PSMILES represents the boundaries of the repeat unit with * and the atoms within the repeat unit using traditional SMILES notation. PC uses a weighted PSMILES (wPSMILES) representation that incorporates repeat unit molar ratios and connectivity to represent copolymers.[18]

To enable property prediction, each of these models convert their polymer inputs into model-specific representations. The language models process PSMILES as tokenized input and predict properties using their pretrained language representations. The graph-based models, PC and PPG, process the molecular graph represented by the PSMILES string, with PC additionally encoding weighted connectivity. The descriptor-based models calculate either machine-learned descriptors, in the case of polyBERT, or expert-designed descriptors, in the case of RDKit.[35] Because the same periodic structure can be represented by distinct PSMILES strings depending on the selected repeat unit, we evaluated prediction consistency across structurally equivalent encodings (Table S2). In general, the descriptor-based models (RF and polyBERT) showed less variance in output predictions based on the specific position of the repeat-unit boundary of the polymer.

Despite their widespread use, these approaches have not been directly compared across a common and open polymer benchmark.[3] polyBERT has been trained and tested on private datasets containing both homopolymers and copolymers;[27] PPG has demonstrated strong prediction performance for several homopolymer properties;[36] and PC has been trained and tested on DFT-calculated copolymer properties.[18,37]

**Table 2**. The models we evaluated on PolyBench26 alongside the user input format, vectorization scheme, and model architecture.

| Model name | Polymer input format | Vectorization scheme | Model architecture |
| --- | --- | --- | --- |
| GPT-4o[32] | PSMILES | Tokens | Transformer based |
| GPT-o3-mini | PSMILES | | |
| Polymer Periodic Graph (PPG)[36] | PSMILES | Graph with periodic edges | Graph Neural Network (GNN), Feedforward Neural Network (FNN) |
| Polymer Chemprop (PC)[18] | Weighted PSMILES (wPSMILES) | Graph with periodic and weighted edges | |
| polyBERT[27] | PSMILES | polyBERT fingerprint | FNN |
| Random Forest (RF) | PSMILES | RDKit descriptors[35] | Random Forest (RF) |

## Task 1: Graph-based models provide the strongest in-distribution property predictions

For Task 1, our aim is to assess the performance of the various models when making in-distribution predictions of polymer properties based on the chemical structure of the repeat unit. We accomplish this aim by randomly partitioning each dataset by unique PSMILES and performing five-fold cross-validation. The resulting test sets are in-distribution with respect to the monomer chemical structures due to our random partitioning. The test sets are also in-distribution with respect to the polymer architecture: models trained on homopolymers are evaluated exclusively on homopolymers and models trained on copolymers are evaluated exclusively on copolymers.

As an initial means to compare the various model architectures, we assessed the performance of the models in Table 2 on homopolymer properties derived from experimental measurements (glass transition temperatures from PolyMetriX) and density functional theory (DFT) calculations (electron affinities from PolyVERSE). We evaluated GPT-4o and GPT-o3-mini as representative large language models, using PSMILES as the text-based representation for the polymers. In our testing, LLM property prediction using

PSMILES outperformed representing polymers by their English name (Table S3). Table 3 summarizes predictive performance for the GPT models alongside polymer-specialized models trained using five-fold cross-validation on the homopolymer datasets, with standard deviations reported across folds. GPT model predictions were generated once per polymer.

**Table 3.** Homopolymer property prediction performance across all model architectures. GPT-model predictions were generated once per polymer using zero-shot prompts, whereas all other results give the test set RMSE using five-fold cross validation. Errors correspond to the standard deviation of the performance across the five folds.

| Model | Electron Affinity (DFT) RMSE (eV) | $T_g$ (Experimental) RMSE (°C) |
| --- | --- | --- |
| GPT-4o | 1.20 | 73.5 |
| GPT-o3-mini | 0.95 | 70.4 |
| PC | **0.29 ± .05** | **34.7 ± 0.6** |
| PPG | **0.29 ± .06** | 36.7 ± 0.8 |
| polyBERT | 0.39 ± .05 | 37.9 ± 1.8 |
| RF | 0.43 ± .06 | 39.7 ± 1.5 |

We find that GPT-o3-mini, a reasoning model, produced lower RMSEs than GPT-4o, although both GPT models exhibited significantly worse performance than the polymer-specialized models under this zero-shot prompting setting. Fine-tuning LLM models on polymer-specific data may improve their performance,[38] but the development of polymer-specific fine-tuning strategies has yet to be fully realized and is beyond the scope of this work.

As a result of the significantly lower performance of the language models under the evaluated prompting conditions, subsequent analyses focused on the polymer-specific models, PC, PPG, polyBERT, and RF. We evaluated these polymer-specific models for in-distribution property prediction on all 14 homopolymer and alternating-copolymer datasets in PolyBench26 (see Table 1). Figure 2 illustrates that among the evaluated methods, graph-based models (PC and PPG) provided the lowest errors for both homopolymers (Figure 2a) and alternating copolymers (Figure 2b). The absolute RMSEs are shown in Table S1.

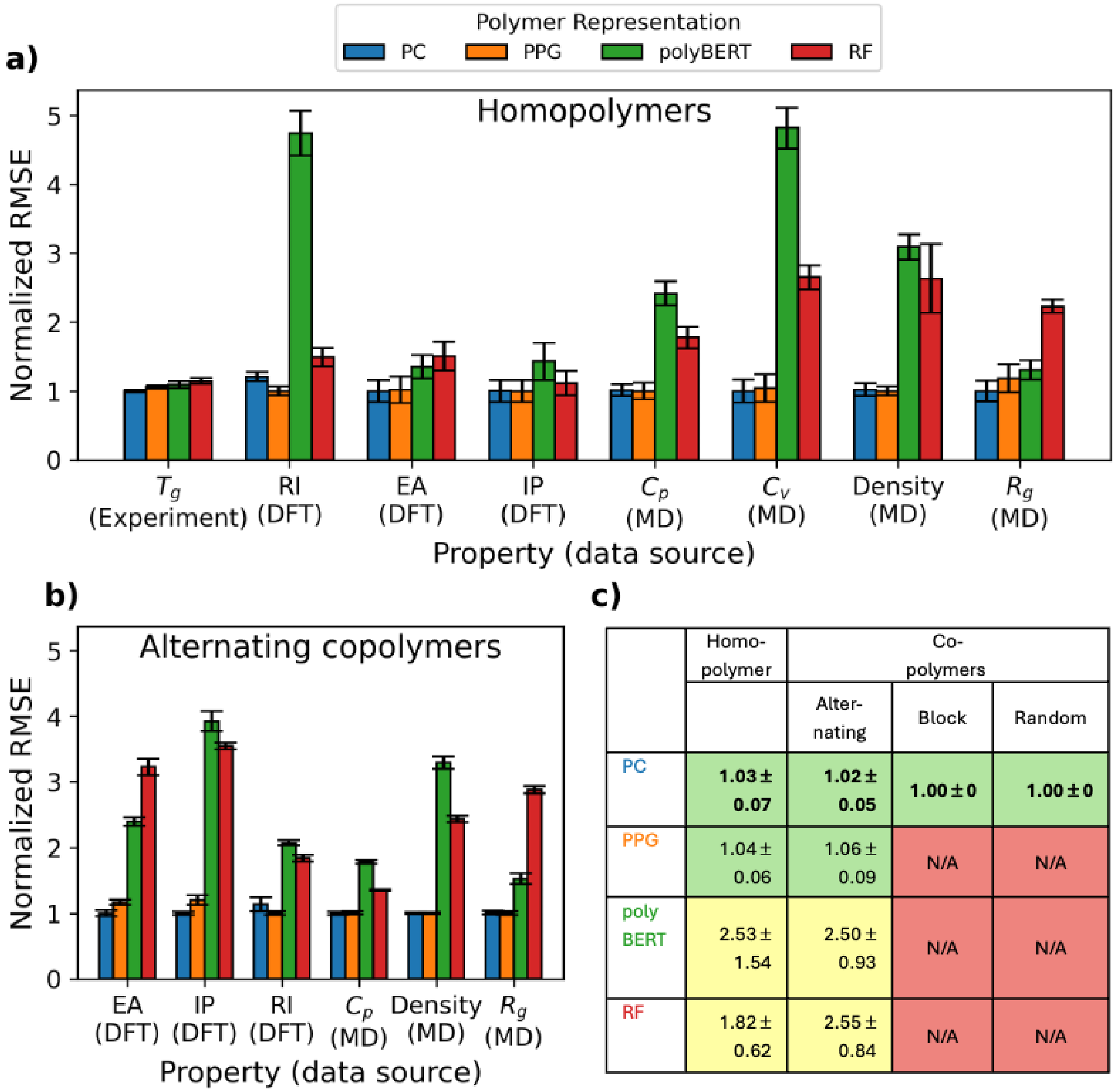

| | Homo-polymer | Co-polymers | | |
|---|---|---|---|---|
| | | Alter-nating | Block | Random |
| PC | **1.03 ± 0.07** | **1.02 ± 0.05** | **1.00 ± 0** | **1.00 ± 0** |
| PPG | 1.04 ± 0.06 | 1.06 ± 0.09 | N/A | N/A |
| poly BERT | 2.53 ± 1.54 | 2.50 ± 0.93 | N/A | N/A |
| RF | 1.82 ± 0.62 | 2.55 ± 0.84 | N/A | N/A |

**Figure 2.** Prediction performance as measured by normalized RMSE across multiple physical properties for a) homopolymers and b) alternating copolymers. RMSE values were normalized by the minimum RMSE of the target property across all models, and error bars indicate one standard deviation across five cross-validation folds. c) Mean normalized RMSE (± standard deviation) for predicted properties of homopolymer and copolymer architectures across model architectures. Values are grouped by polymer architecture. Green cells indicate models whose mean normalized RMSEs are not statistically distinguishable from the best-performing model (bold) in the same column. Yellow cells indicate significantly higher mean normalized RMSE than the best-performing model in the same column. Red cells indicate polymer architectures that the model cannot natively represent.

In Figure 2c, we summarize model performance as the mean normalized RMSE for properties grouped by polymer architecture. The reported uncertainty reflects variability across the evaluated properties. Results highlighted in yellow have statistically higher mean normalized RMSE than the best-performing model in the same column according to a paired Student's *t*-test ($p < 0.01$).

Of the evaluated model architectures, PC is the only model that natively represents block and random copolymers through wPSMILES. The other models are not directly compatible with these polymer architectures under the representations used here, and their results are therefore reported as not applicable (N/A) in Figure 2c. For both homopolymer and alternating copolymer prediction tasks, the graph-based models (PC and PPG) performed similarly and provided the lowest RMSEs among the evaluated methods. As a result, we conclude from this Task that polymer property prediction models based on a graph architecture consistently show the best performance on in-distribution prediction tasks.

## Task 2: Model performance scales consistently with training-set size across polymer properties and polymer architectures

In Task 2, we explore how model performance is impacted by the size of the training set. This Task is particularly important for polymer informatics since data scarcity may degrade the performance of models when trained on such small datasets. To evaluate model performance with limited datasets, we held the validation and test sets fixed while varying the training-set size from 300 datapoints up to the full available training set. We tested the models on subsets of the largest homopolymer dataset, glass-transition temperature, and each of the OPoly26 datasets (Figures 3, S1, and S2).

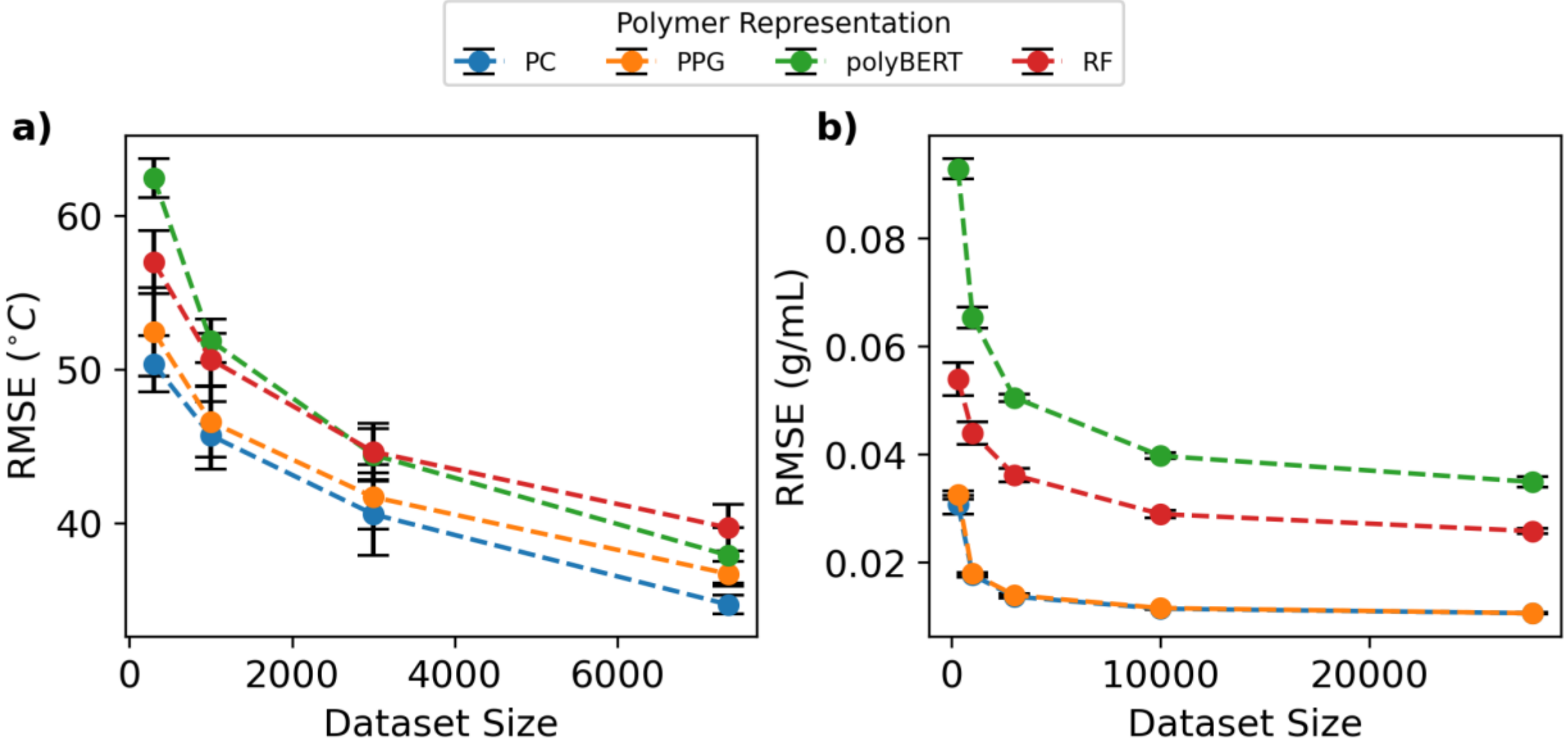

**Figure 3.** The dependence of prediction error on training dataset size for representative homopolymer and alternating copolymer properties, with (a) $T_g$ shown for homopolymers and (b) density shown for alternating copolymers. Although only one property is shown for each polymer class, all evaluated quantities exhibit qualitatively similar dataset size scaling behavior (Figures S1 and S2).

In Figure 3, we show the results obtained from Task 2 for two representative datasets: homopolymer $T_g$ and alternating copolymer densities (see Figures S1 and S2 for full results). Similar to the full-data experiments of Task 1, we find that graph-based models exhibited lower RMSEs across the evaluated properties and training-set sizes. Additionally, we find that the relative performance of the models remained similar across the evaluated dataset sizes. The scaling behavior observed for the MD-derived properties was consistent across the evaluated quantities, shown in Figures S1 and S2. For the single DFT-derived refractive index (Figure S2, panel d), the results were consistent with the DFT-computed polymer properties of EA and IP results reported by Aldeghi and Coley.[18] Specifically, graph-based models outperformed descriptor-based models only after the training set exceeded approximately 1,000 datapoints. For experimental and MD-derived properties, in contrast, our results show graph-based models achieve the lowest RMSE for every training-set size evaluated, including the smallest subset of 300 unique polymers. Although graph-based models exhibited the lowest RMSEs in both the high- and low-data regimes, they also required the longest training times across the evaluated datasets (Tables S4 and S5). Descriptor-based models may therefore be favored in applications where computational efficiency outweighs the improvements in predictive accuracy.

## Task 3: Graph-based models remain robust as repeat-unit complexity increases

In Task 3, we evaluate the effect of repeat-unit complexity on model performance. This analysis is particularly relevant for polymer representations that encode the entire repeat unit with global features (such as the RDKit descriptors and polyBERT), because increased repeat-unit complexity may make it more difficult for these representations to retain relevant local and relationship information.

To assess model behavior as the complexity of repeat units increase, we evaluated prediction error on the DFT-calculated refractive index dataset using copolymers with repeat units containing five to ten chemically distinct monomers arranged in an alternating sequence. Here, repeat-unit complexity is defined by the number of distinct monomers within the repeating sequence. Representative repeat unit structures are shown in Figure 4b. As shown in Figure 4a, the RMSE of PC and PPG remained low across this complexity range, whereas the RMSE of descriptor-based models polyBERT and RF increased with the number of monomer units in each repeat unit. The observed divergence indicates that graph-based representations better preserve relational chemical information as the repeat unit becomes more complex. By contrast, the evaluated descriptor-based representations appeared less able to distinguish the structural information needed for accurate prediction in this regime.

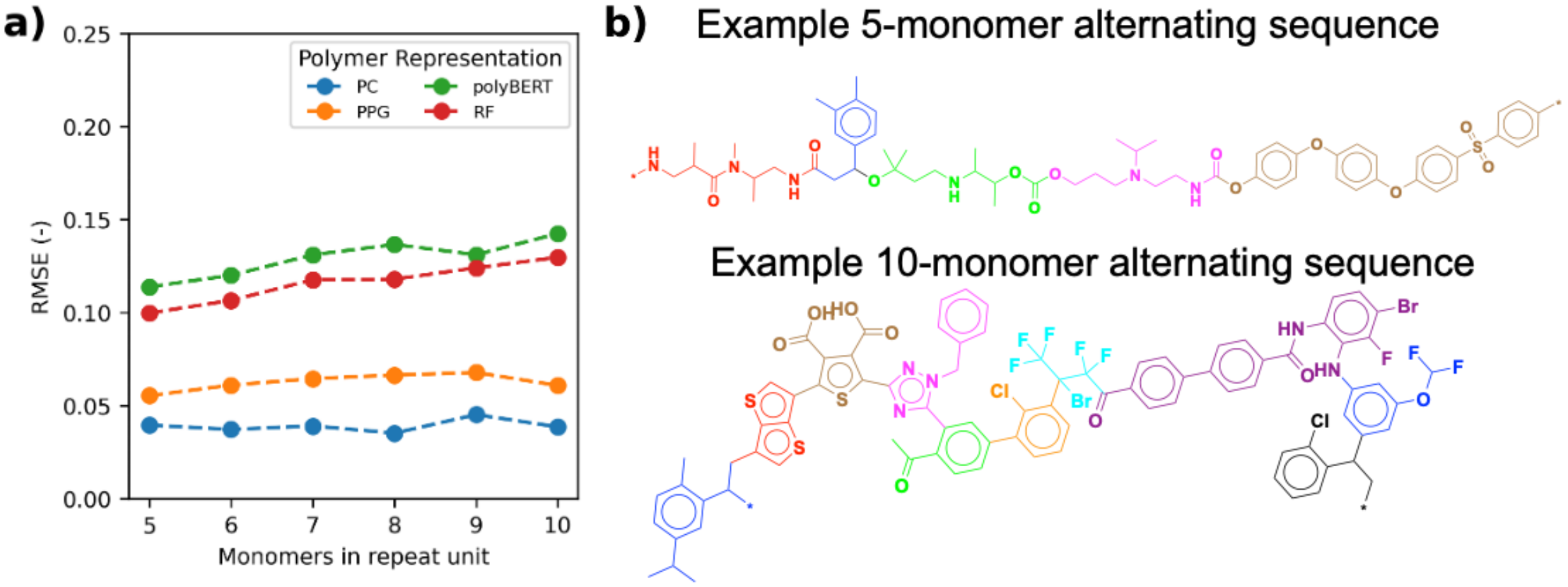

**Figure 4.** a) Prediction error for alternating copolymers' refractive index as a function of the number of distinct monomers in the repeat unit for different polymer representations. Points show mean RMSE, and dashed lines are provided to guide the eye. Graph-based message-passing representations (PC and PPG) maintain consistently low error with increasing repeat unit complexity, while descriptor-based models (polyBERT and RF) increase with repeat unit complexity. b) Example alternating polymer repeat units with each monomer drawn in a different color, showing the increasing complexity of the chemical structures.

## Task 4: Weighted graph representations support transfer across held-out polymer architectures

In Task 4, we evaluate model generalization across polymer architectures by testing models on polymer architectures that were not seen in training. Because polymeric materials exhibit substantial structural diversity, it is desirable for models trained on one polymer architecture, such as alternating copolymers, to generalize to other polymer architectures, such as block or random copolymers. We therefore performed a polymer architecture-held-out evaluation by first training the models on a combined dataset of homopolymer and alternating-copolymer electron-affinity values. We then evaluated them on test sets spanning four polymer architectures: homopolymers and alternating copolymers, which were represented during training, and random and block copolymers, which were held out from training. Importantly, while we excluded random and block copolymers from training, we retained their constituent repeat units in the homopolymer and alternating copolymer training data.

The experimental results shown in Figure 5 evaluate transfer across polymer architecture rather than extrapolation to entirely unseen repeat-unit chemistries. Consistent with Figure 2, the graph-based models achieved lower prediction errors than polyBERT and RF for alternating copolymers. Among the evaluated models, only PC natively represents random and block copolymers through wPSMILES. For the other models, we approximated copolymer electron affinity as the molar-ratio-weighted average of predictions for the individual repeat units.

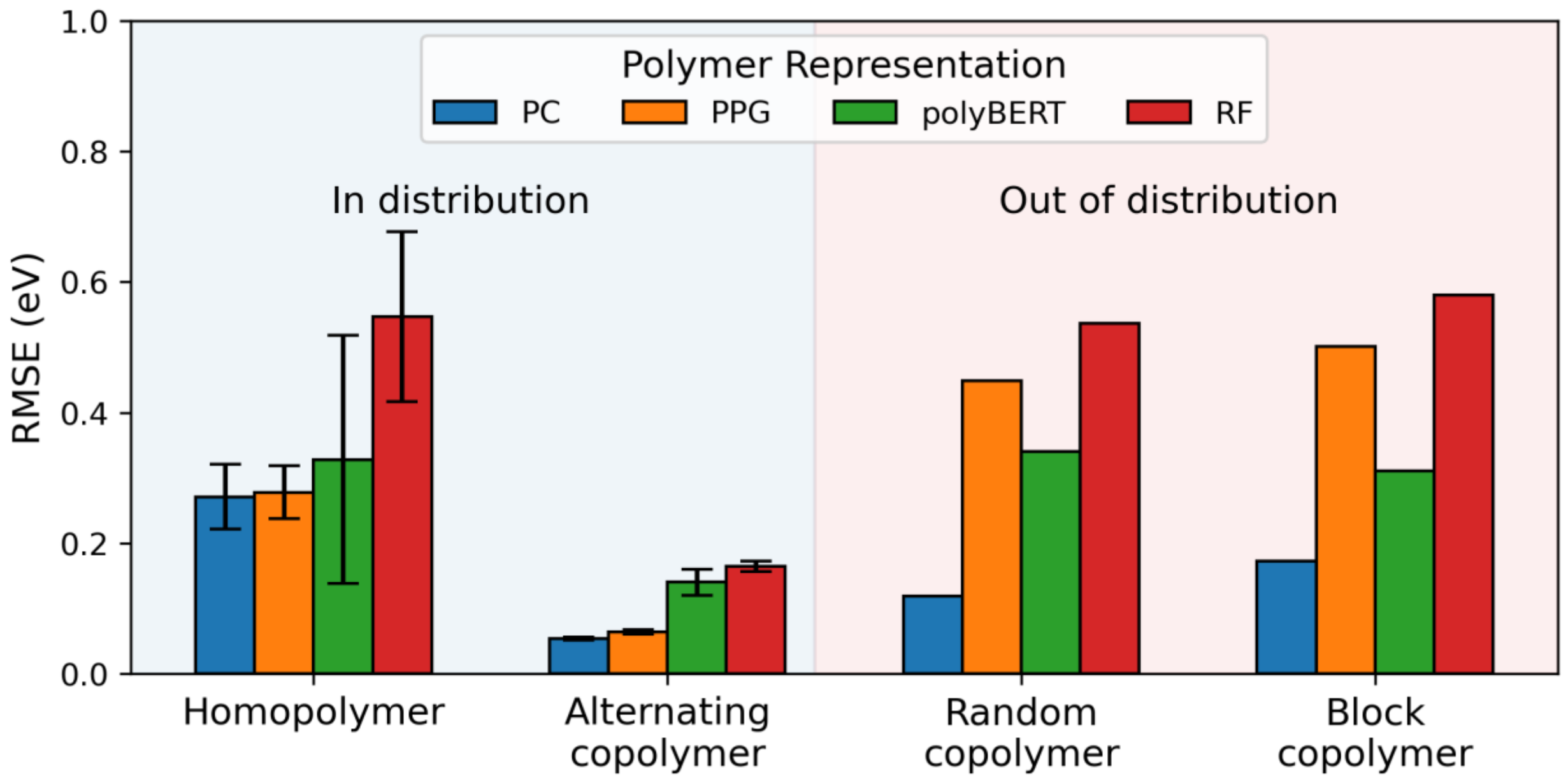

**Figure 5**. Electron affinity (EA) prediction performance for different polymer architectures using models trained on a combined dataset of homopolymers and alternating copolymers. Results are shown for in-distribution polymer architectures (homopolymers and alternating copolymers) and out-of-distribution polymer architectures (random and block copolymers). For in-distribution data, the bars show the mean RMSE across 5 cross-validation folds, with error bars indicating one standard deviation. For out-of-distribution random and block copolymers, the bars show the RMSE of a single test set and all pairwise differences between model RMSEs are statistically significant ($p < 0.001$), as determined by a two-sided Wilcoxon ranked sign test.

PC showed the strongest generalization performance on the held-out random- and block-copolymer test sets. We attribute the differences in performance to the neural network architectural design choices underlying the evaluated representations. By explicitly encoding bond weights, PC preserves information about monomer composition and connectivity in a form that remains applicable when the polymer architecture changes. In contrast, models lacking an explicit representation of polymer architecture are effectively constrained to predictions based on weighted combinations of predictions for the constituent repeat-units. These approaches produced RMSEs approximately twofold (polyBERT) to threefold (PPG and RF) higher than PC on the held-out polymer architectures.

# Discussion

We present Polymer Benchmark 2026 (PolyBench26), an open database comprising nearly 250,000 polymer-property datapoints across eight physical properties and four polymer architectures. Through four standardized evaluation tasks, in-distribution prediction, dataset-size scaling, repeat-unit complexity, and generalization across held-out polymer architectures, PolyBench26 enables systematic comparison of polymer-property prediction methods. To promote reproducibility and broad community adoption, we will release the complete dataset accompanying this study. We subsequently plan to generate an independent, hidden test set using new molecular-dynamics simulations to support a Hugging Face leaderboard for unbiased evaluation of new machine learning architectures for polymer property prediction.

Among the evaluated methods, graph-based models provided the lowest errors for homopolymers and alternating copolymers, maintained their relative performance as the training-set size decreased, and remained robust as repeat-unit complexity increased. PC explicitly encodes monomer connectivity and composition through weighted graph edges, performs competitively for all tasks, and was the only evaluated implementation that natively represented all polymer architectures considered here. These results support broader adoption of PC and wPSMILES as a method for polymer-property prediction, particularly when copolymer composition and architecture are central to the application. To facilitate their use, we will also release helper functions for converting between PSMILES and wPSMILES representations of homopolymers, and alternating, random, and block copolymers. More generally, polymer-specific representations that explicitly encode connectivity can support transfer across polymer architectures.

Despite these advances, reliable surrogate modeling for copolymers remains challenging. The polymer architectures considered here represent relatively tractable cases, whereas more difficult systems include network polymers, block copolymers with multiple glass-transition temperatures, and multiblock, gradient, and graft copolymers. Addressing these systems will require richer representations and expanded, high-quality datasets that capture their structural complexity and experimentally relevant behavior. PolyBench26 provides a foundation for developing and testing such methods, but continued progress will depend on broadening benchmark coverage to a wider range of polymer architectures.

# Methods

## Data Sources and Curation

### Polymer Property Datasets

The Polymer Benchmark 2026 (PolyBench26) aggregates polymer property datasets spanning experimental measurements, density functional theory (DFT) calculations, and molecular dynamics (MD) simulations. The benchmark includes homopolymers and copolymers across alternating, random, and block architectures.

DFT-calculated electron affinity (EA) and ionization potential (IP) data for homopolymers were obtained from polyVERSE.[22,39] Copolymer EA and IP values (42,966 for each property) were taken from the VIPEA dataset by Aldeghi and Coley.[18] Experimental glass transition temperature ($T_g$) data comes from PolyMetriX.[31] We extracted an additional 155,930 property datapoints from existing OPoly26 MD trajectories.[16]

Across all homopolymer datasets, the benchmark contains 17,003 individual property measurements corresponding to 9,905 unique homopolymer repeat units. Copolymer datasets include 232,959 polymer–property datapoints and 80,099 unique copolymers.

For consistent evaluation across models, we converted weighted PSMILES from VIPEA to traditional PSMILES for models other than PC. Conversely, traditional PSMILES from PolyMetriX and polyVERSE were converted to wPSMILES for PC. Both representations were retained for the simulated polymers.

### OPoly26 Dataset

The molecular dynamics (MD) trajectories were obtained from the Open Polymers 2026 (OPoly26) dataset,[16] which comprises all-atom simulations of amorphous polymer systems generated using LAMMPS and RadonPy.[40,41] The original trajectory-generation workflow, including the modified GAFF2 force field, periodic boundary conditions, and multistep equilibration procedure is described in detail in the cited OPoly26 and RadonPy references.[16,42–44] The OPoly26 dataset contains MD trajectories of 94,436 unique amorphous polymer simulation cells. We only include the trajectories that: i) contain approximately 5,000 total atoms, ii) do not contain 'ring spears' (a bond that penetrates through the face of a ring), and iii) do not contain solvent molecules or ions, reducing the number of trajectories used to 38,933. The calculation procedures for each individual property are detailed below.

#### *Refractive Index*

The Lorentz-Lorenz equation is used to calculated refractive index, $n$:[41]

$$\frac{n^2 - 1}{n^2 + 2} = \frac{4\pi\rho}{3M_W}\alpha_{polar}$$

where $M_W$ is the molecular weight, $\rho$ is the density, and $\alpha_{polar}$ is the isotropic dipole polarizability of the repeat unit.

### *Density*

The density, $\rho$, in an *NpT* simulation was calculated using the mass *m* and volume *V* of the system as follows:

$$\rho = \frac{m}{\langle V \rangle}$$

where the angular brackets represent time averaging.

### *Radius of gyration*

The radius of gyration, $R_g$, was calculated using the following equation:

$$R_g = \sqrt{\frac{1}{N}\sum_{k=1}^{N}(\boldsymbol{r}_k - \boldsymbol{r}_{mean})^2}$$

where $\mathbf{r}_k$ is the position of the *k*-th repeat unit, and $\mathbf{r}_{mean}$ denotes the mean position of the repeat units of the polymer chain.

### *Heat capacity*

The specific heat capacity at constant pressure $C_p$ was calculated from fluctuations in the enthalpy, *H*:

$$C_p = \frac{\langle \delta H^2 \rangle}{k_B T^2 m}$$

where $k_B$ is the Boltzmann constant and *T* is the temperature. The enthalpy was calculated using the constant pressure of 1 atm because the calculated pressure value in *NpT* simulations has a significant fluctuation. Classical heat capacity is known to be problematic due to strong quantum nuclear effects. The classical value (governed by Dulong-Petit law, $C_v = 3RN$), is effectively a ceiling for $C_v$, since classical MD considers all vibrational modes to be active, regardless of temperature. This $C_p$ calculation method results in computational values of $C_p$ consistently higher than experimentally measured values reported on PolyInfo (Figure S5).[34,41,45]

To more accurately predict $C_v$, we first start from the equilibrated NPT calculation (300 K, 1 atm) and perform an initial thermalization in NVT. Specifically, we re-assign velocities from a Maxwell-Boltzmann distribution, and run NVT (with a Nosé-Hoover thermostat) calculations at 300 K, with a time constant of 100 fs. After 100 picoseconds of NVT dynamics, we run 100 ps of NVE dynamics; both steps use an initial timestep of 1 femtosecond. Then, we perform a final production or sampling period, with NVE dynamics and a 0.5 femtosecond timestep. Atomic coordinates are saved in double-precision, ingested by TRAVIS, and used to compute $C_v$.

We opt to compute and numerically integrate the phonon spectrum, as described below. Within this workflow, the phonon spectrum is calculated via the Power Spectra module available within the software package TRAVIS.[46] TRAVIS does finite differencing of atomic positions from a classical MD trajectory to interpolate velocities, which are then used to compute a global velocity autocorrelation function (VACF), and its Fourier transform, $P(\tilde{\nu})$. To compute the normalized phonon density of states, we use

$$g(\tilde{\nu}) = N_{\text{modes}} \frac{P(\tilde{\nu})}{\int_0^\infty P(\tilde{\nu}) d\tilde{\nu}}$$

where the total number of vibrational modes is $N_{\text{modes}} \approx 3N$. The resulting spectra, reported by TRAVIS, have a resolution of approximately $\Delta\tilde{\nu} \approx 2$ cm$^{-1}$, and $\tilde{\nu}_{\text{max}} \approx 5000$ cm$^{-1}$. Finally, the normalized phonon density of states, $g(\tilde{\nu})$, is substituted into the following expression for $C_v$:

$$C_V(T) = k_b \int_0^\infty (\beta h \tilde{\nu} c)^2 \frac{e^{\beta h \tilde{\nu} c}}{(e^{\beta h \tilde{\nu} c} - 1)^2} g(\tilde{\nu}) d\tilde{\nu}.$$

This expression accounts for the Bose-Einstein statistics of phonons and quantization; here $\tilde{\nu}$ is wavenumber, $\beta = (k_B T)^{-1}$, and the parameters $c$, $h$, and $k_B$ are standard physical constants (speed of light, Planck's constant, and Boltzmann's constant), and $T \approx 300$ K. As a first approximation for this model, we opted to compare $C_v$ values to experimental $C_p$ values, on the basis that the correction should be modest. However, we note that this approximation will lead to a systematic under-prediction relative to $C_p$ (Figure S6).

Additionally, atomic positions are recorded in double precision, to ensure accurate finite differencing within TRAVIS. To maximize the speed of the Fast Fourier Transform (FFT) of the VACF, we restrict the production trajectory length to a power of 2, selecting a window size of $2^{15} = 32768$ frames. Positions are sampled every 0.5 fs, which amounts to a NVE production trajectory of approximately 16.4 ps in length. However, we tested sampling periods of various lengths up to approximately 66 ps. Prior to this NVE production run, each system is prepared at 300 K and 1 atm, following the procedure outlined earlier in the manuscript.

## Polymer Representations

### SMILES and PSMILES Processing

Homopolymer PSMILES strings were converted to weighted PSMILES (wPSMILES) format by replacing the first and second asterisks with [*:1] and [*:2], respectively, and appending |1|<1-2:1:1<1-1:0:0<2-2:0:0 to encode homopolymer connectivity.

Alternating copolymer wPSMILES strings from prior work were converted to standard PSMILES by replacing numbered connection points with asterisks and concatenating repeat units while preserving valid SMILES syntax.

For random and block copolymers, stoichiometric ratios were encoded explicitly in the weighted message-passing framework (PC model). For models lacking explicit copolymer architecture support, copolymer predictions were approximated as molar-ratio-weighted averages of individual repeat-unit predictions.

PSMILES representations from polyVERSE and PolyMetriX were retained without additional structural standardization, such as canonicalization. Each model interprets the atoms at the boundary of the repeat unit differently and the choice of where to place the edges of the repeat unit window can affect the polymer representation and predictions (Table S2).

## Fingerprinting

### polyBERT Fingerprints

We use the polyBERT embedding model downloaded from Huggingface (accessed November 6th, 2025) to convert the PSMILES strings to their 600-dimensional fingerprints prior to training the property prediction models. To prepare alternating copolymers for training, we use the alternating_copolymer() function within the PSMILES Python package, as recommended by the authors.

### RDKit Descriptors

RDKit[35] (version 2025.9.1) was used to compute 2D molecular descriptors for repeat units. Descriptors with missing values across the dataset were removed. Missing values were replaced with the median value within each dataset.

## Machine Learning Models

### Graph-Based Message Passing Models

PC and PPG models were built on top of Chemprop (version 1.4.0)[47,48] and used directly from the authors' repositories with minimal modification.[18,36] Models were trained using directed message passing neural networks with a fixed seed, 5-fold cross-validation, and 50 epochs. All other hyperparameters were set to their default values.

### polyBERT + Feedforward Neural Network

polyBERT fingerprints were used as input to a feedforward neural network (FNN) matching the hyperparameters of the FNN used in the graph-based message passing models. Briefly, this FNN had two hidden layers with 300 neurons each, an initial learning rate of $10^{-4}$, and a batch size of 50. The full configuration file can be found on the Github repository associated with this work.

### Descriptors + Random Forest Baseline

### Training Protocol

All models were trained using 5-fold cross-validation. For each fold, data were de-duplicated on PSMILES before being randomly split into 80% training, 10% validation, and 10% test sets.

For transfer experiments, models were trained on combined homopolymer and alternating copolymer datasets and evaluated on held-out in-distribution (homopolymer and alternating copolymer) and out-of-distribution (random and block copolymer) test sets. Out-of-distribution copolymer architectures were excluded entirely from training.

Dataset size experiments were conducted by training models on randomly sampled subsets of increasing size, while keeping the same fixed validation set and test set.

### GPT Evaluation Protocol

Large language models (GPT-4o and GPT-o3-mini) were accessed via API. Each polymer was queried once per prompting strategy. Predicted numeric values were extracted via structured parsing of the model output. No fine-tuning was performed. Example prompts and a comparison between PSMILES and English language representation are available in the Supporting Information (Tables S4 and S5).

## Evaluation Metrics

### Normalized RMSE

To enable comparison across properties with different units and magnitudes, RMSE values were normalized by the minimum RMSE of the target property across all models. Normalization was performed separately for each property.

### Statistical Analysis

Statistical significance between models was evaluated using paired two-sided Student's t-tests across cross-validation folds. For out-of-distribution single-test evaluations, pairwise comparisons were performed using two-sided Wilcoxon signed-rank tests.[49]

## Data and Code Availability

All datasets and splits, preprocessing scripts, model training code, and evaluation routines are publicly available at: https://github.com/rlearsch/PolymerBenchmark2026. We plan to host an online leaderboard for model performance based on an independent, hidden test generated using molecular-dynamics simulations, to support unbiased evaluation of new machine learning architectures for polymer property prediction. The leaderboard will be available via the Github repository.

## Acknowledgements

This work was produced under the auspices of the U.S. Department of Energy by Lawrence Livermore National Laboratory under Contract DE-AC52-07NA27344. E.R.A., A. M. H., and R. W. L. acknowledge support from the Laboratory Directed Research and Development Program of Lawrence Livermore National Laboratory, project number LDRD 24-SI-008. N.L. acknowledges support from the Laboratory Directed Research and Development Program of Lawrence Livermore National Laboratory, project number LDRD 23-ERD-030. The PolymerBench26 data and code are released under LLNL Release LLNL-DATA-2017526 and paper release LLNL-JRNL-2024204.

# Supporting Information

## Modeling results for all properties and model architectures

**Table S1**. RMSE and standard deviations from 5-fold cross validation for all specialized models and polymer datasets

| Property | Source | PPG RMSE | PC RMSE | polyBERT RMSE | RF RMSE |
|---|---|---|---|---|---|
| Homopolymers | | | | | |
| Electron Affinity (eV) | DFT | 0.291 ± 0.055 | **0.285 ± 0.045** | 0.385 ± 0.049 | 0.429 ± 0.059 |
| Ionization potential (eV) | DFT | **0.425 ± 0.069** | 0.426 ± 0.067 | 0.608 ± 0.115 | 0.473 ± 0.075 |
| Glass transition temperature (°C) | Experimental | **36.7** ± 0.8 | **34.7** ± 0.6 | 37.9 ± 1.8 | 39.7 ± 1.5 |
| Specific heat capacity ($C_p$) (J/kg K) | MD | **85.2 ± 10.2** | 86.3 ± 7.4 | 205.8 ± 15 | 151.4 ± 13.6 |
| Specific heat capacity ($C_v$) (J/g K) | MD | 0.013 ± 0.002 | **0.012 ± 0.002** | 0.058 ± 0.004 | 0.032 ± 0.002 |
| Density (g/mL) | MD | **0.025 ± 0.002** | **0.025 ± 0.002** | 0.076 ± 0.005 | 0.065 ± 0.012 |
| Refractive index (n) (-) | DFT | **0.015 ± 0.001** | 0.019 ± 0.001 | 0.073 ± 0.005 | 0.023 ± 0.002 |
| Radius of gyration (nm) | MD | 3.4 ± 0.6 | **2.8 ± 0.4** | 3.7 ± 0.4 | 6.3 ± 0.3 |
| Alternating copolymers | | | | | |
| Specific heat ($C_p$) (J/kg K) | MD | 67.8 ± 1.51 | **67.6 ± 1.67** | 120.3 ± 1.83 | 91.3 ± 0.89 |
| Density (g/mL) | MD | **0.011 ± 0.000** | **0.011 ± 0.000** | 0.035 ± 0.001 | 0.026 ± 0.001 |
| Refractive index (n) (-) | DFT | **0.055 ± 0.002** | 0.063 ± 0.006 | 0.115 ± 0.002 | 0.102 ± 0.003 |
| Radius of gyration (nm) | MD | **1.22 ± 0.04** | 1.23 ± 0.04 | 1.86 ± 0.10 | 3.52 ± 0.07 |
| Electron affinity (EA) (eV) | DFT | 0.056 ± 0.002 | **0.048 ± 0.002** | 0.115 ± 0.003 | 0.155 ± 0.006 |
| Ionization potential (IP) (eV) | DFT | 0.048 ± 0.003 | **0.040 ± 0.001** | 0.157 ± 0.006 | 0.142 ± 0.002 |

| Random copolymers | | | | | |
|---|---|---|---|---|---|
| Specific heat ($C_p$) (J/kg K) | MD | - | **96 ± 2.6** | - | - |
| Density (g/mL) | MD | - | **0.023 ± 0.001** | - | - |
| Refractive index (n) | DFT | - | **0.07 ± 0.005** | - | - |
| Radius of gyration (nm) | MD | - | **2.05 ± 0.065** | - | - |
| Electron affinity (EA) (eV) | DFT | - | **0.027 ± 0.001** | - | - |
| Ionization potential (IP) (eV) | DFT | - | **0.021 ± 0.001** | - | - |
| Block copolymers | | | | | |
| Electron affinity (EA) (eV) | DFT | - | **0.033 ± 0.002** | - | - |
| Ionization potential (IP) (eV) | DFT | - | **0.024 ± 0.001** | - | - |

## Model sensitivity to different text representation of the same chemical structures

**Table S2.** Comparison of Model Density Predictions for Structurally Equivalent Polymers Represented by Distinct PSMILES Strings.

| Structure | PSMILES | PPG density predictions | PC density predictions | polyBERT density predictions | RF density predictions |
|---|---|---|---|---|---|
| Poly(ethylene glycol) | | | | | |
| | *CCO* | 1.232 | 1.196 | 1.202 | 1.16 |
| | *COC* | 1.170 | 1.196 | 1.202 | 1.36 |
| | *OCC* | 1.232 | 1.196 | 1.202 | 1.16 |
| Poly(ethylene terephthalate) | | | | | |
| | O=C(OCC[*])C1=CC=C(C(O[*])=O)C=C1 | 1.391 | 1.388 | 1.405 | 1.38 |
| | O=C(OCCOC([*])=O)C1=CC=C([*])C=C1 | 1.379 | 1.386 | 1.405 | 1.36 |
| | O=C(OC[*])C1=CC=C(C(OC[*])=O)C=C1 | 1.358 | 1.388 | 1.405 | 1.36 |
| Nylon 6 | | | | | |
| | [*]NCCCCCC([*])=O | 1.146 | 1.137 | 1.126 | 1.19 |
| | [*]CC(=O)NCCCC[*] | 1.163 | 1.149 | 1.126 | 1.22 |
| | [*]NC(=O)CCCCC[*] | 1.162 | 1.149 | 1.126 | 1.15 |
| Polyurethane | | | | | |
| | O=C([*])NC1=CC=C(NC(OCCO[*])=O)C=C1 | 1.380 | 1.417 | 1.314 | 1.35 |
| | O=C(OC[*])NC1=CC=C(NC(OC[*])=O)C=C1 | 1.365 | 1.416 | 1.314 | 1.36 |
| Poly(ethylenimine) | | | | | |
| | *CCN* | 1.182 | 1.146 | 1.163 | 1.16 |
| | *CNC* | 1.157 | 1.146 | 1.163 | 1.39 |
| | *NCC* | 1.182 | 1.146 | 1.163 | 1.62 |
| Poly(dimethyl propane) | | | | | |
| | *CC(C)(C)C* | 0.898 | 0.855 | 0.982 | 0.95 |
| | *CCC(C)(C)(*) | 0.922 | 0.855 | 0.875 | 0.89 |
| | C(C)(C)(*)CC* | 0.922 | 0.855 | 0.875 | 0.89 |

Each PSMILES listed corresponds to an equivalent repeat unit of the same polymer structure; these variations demonstrate the sensitivity of certain representations to differences in textual encoding despite chemical equivalence.

## All dataset size results:

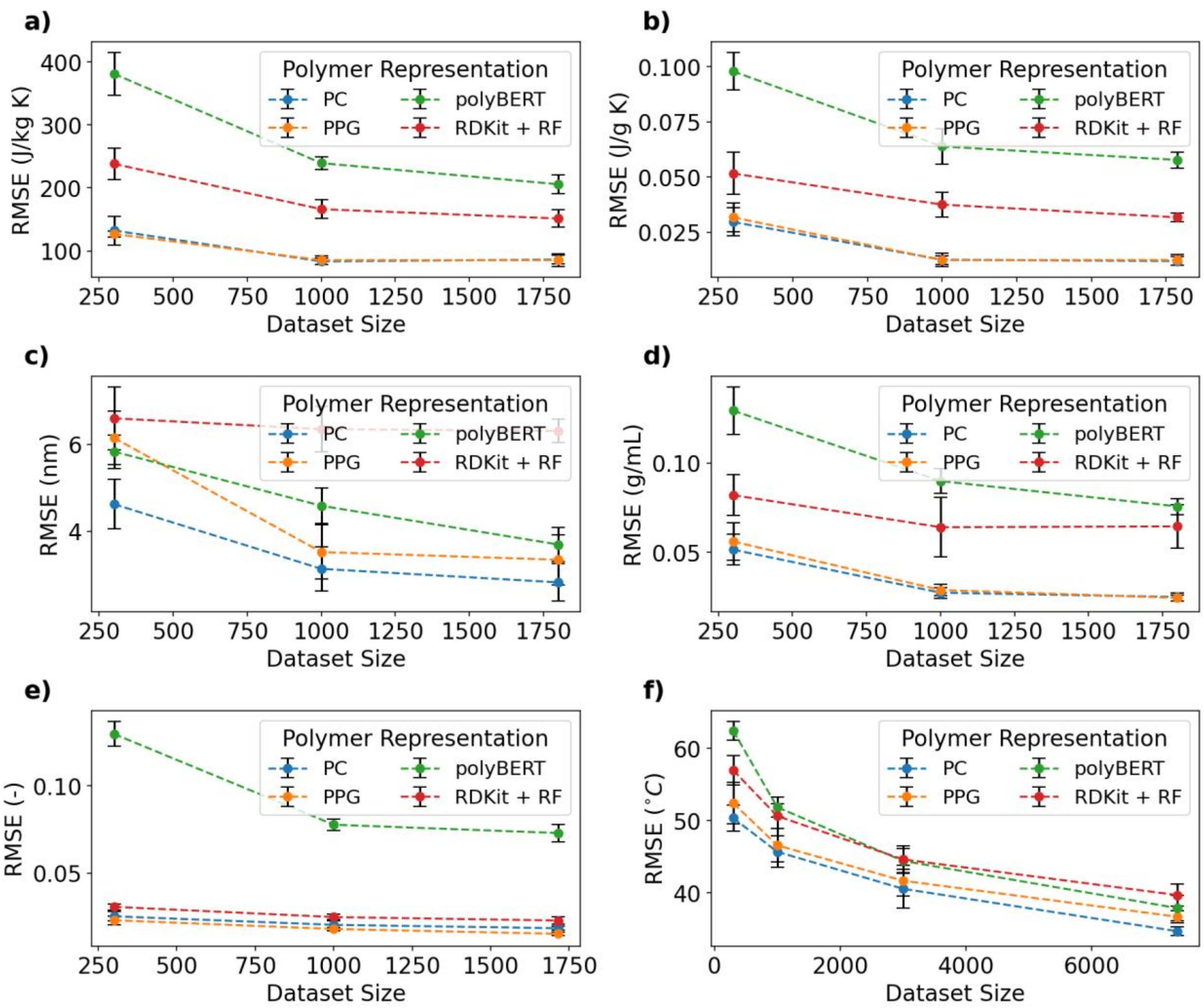

**Figure S1**: The dependence of prediction error on training dataset size for homopolymer properties: a) $C_p$, b) $C_v$, c) $R_g$, d) density, e) refractive index, f) $T_g$. The error bars show the standard deviation in RMSE calculated from 5-fold cross validation.

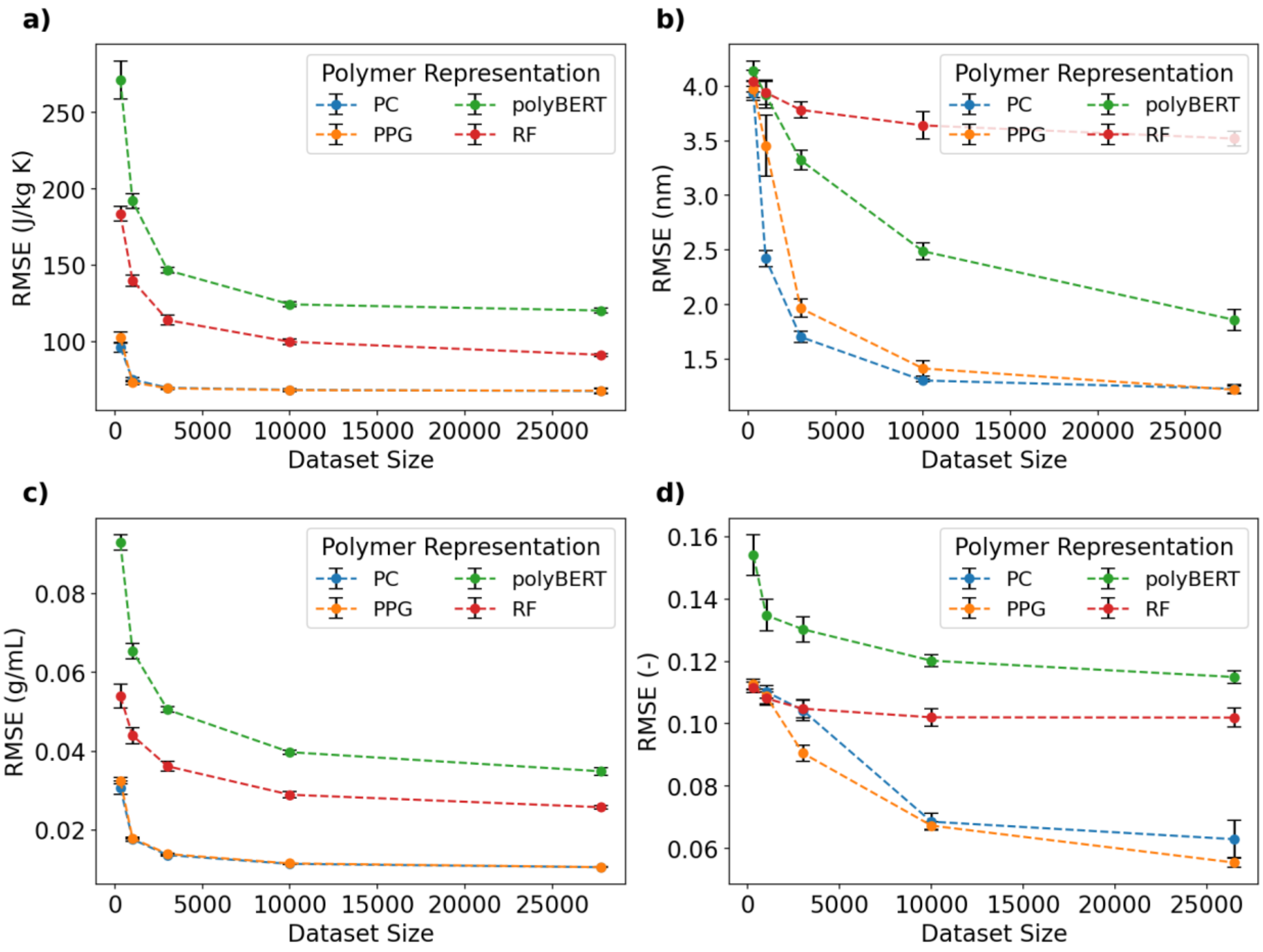

**Figure S2**. The dependence of prediction error on training dataset size for alternating copolymer properties: a) $C_p$, b) $R_g$, c) density, d) refractive index. The error bars show the standard deviation in RMSE calculated from 5-fold cross validation.

## Comparing homopolymer results between experimental measurements and MD measurements

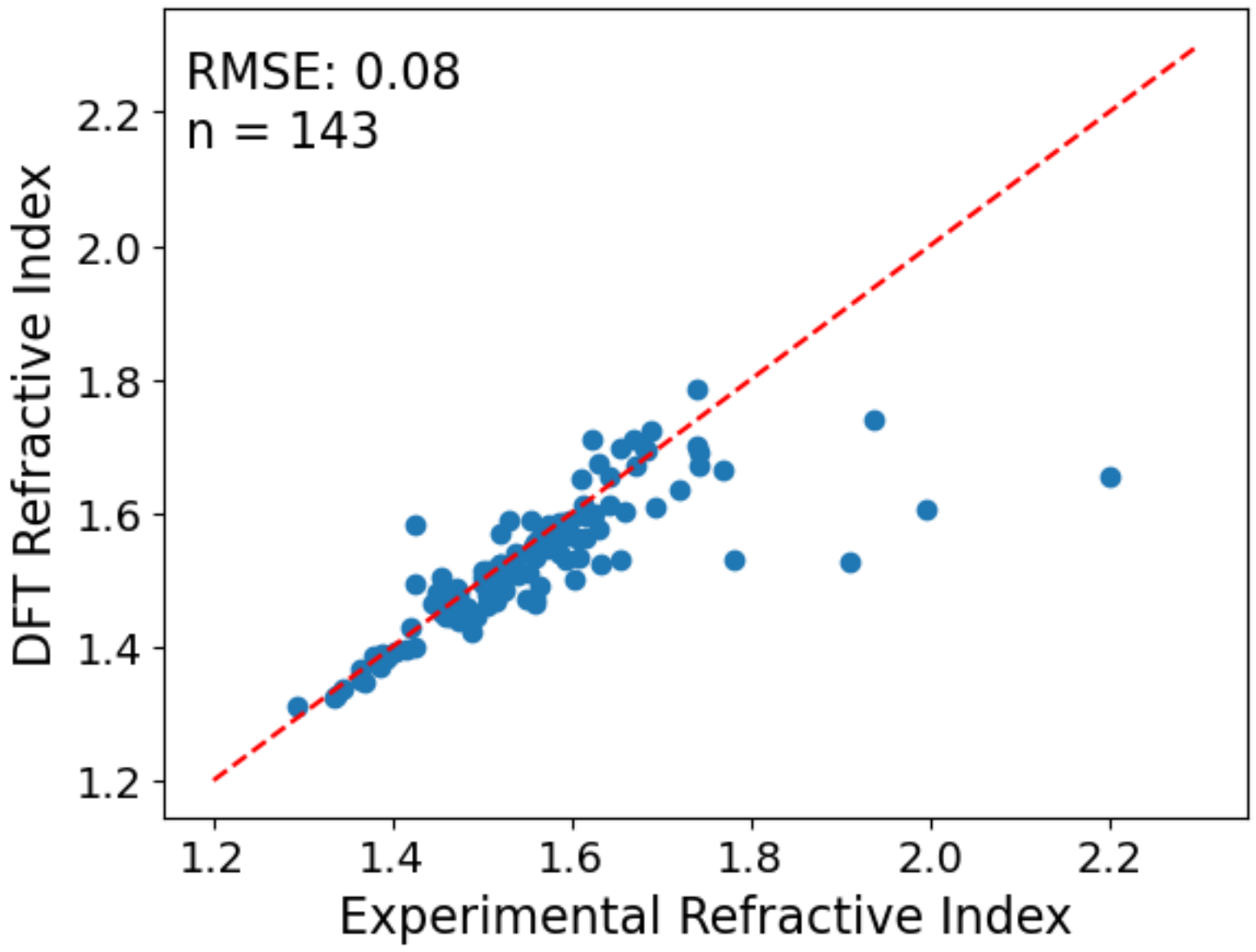

**Figure S3**. Comparison between DFT-calculated refractive index (OPoly26) and experimental refractive index values. $r^2 = 0.63$. The dashed red line indicates y=x.

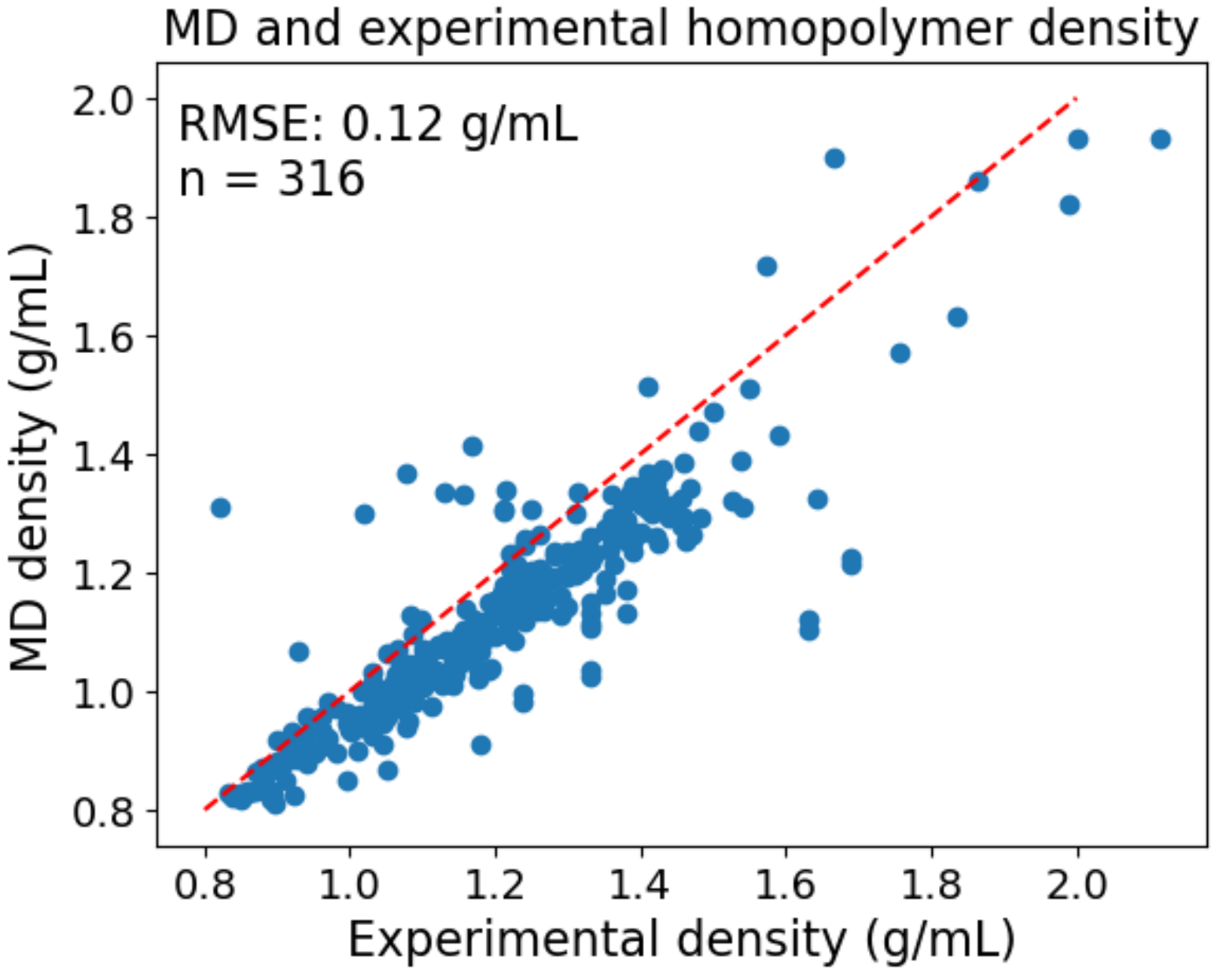

**Figure S4**: Comparison between MD-calculated density (OPoly26) and experimental density values. $r^2 = 0.79$. The dashed red line indicates y=x.

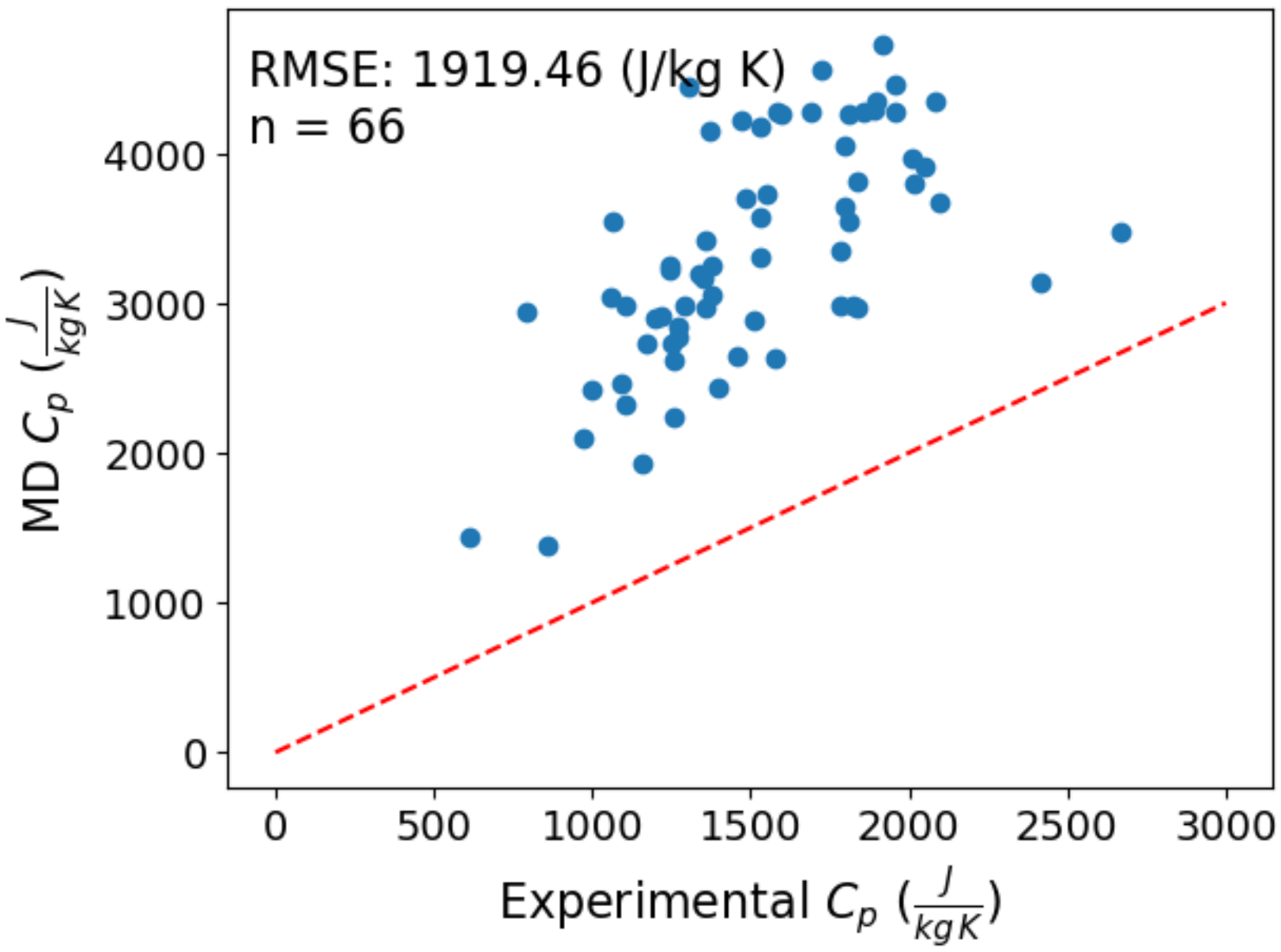

**Figure S5**: Comparison between traditional MD-calculated $C_p$ (OPoly26) and experimental specific heat capacity values from PolyInfo. $r^2 = 0.40$. The dashed red line indicates y=x.

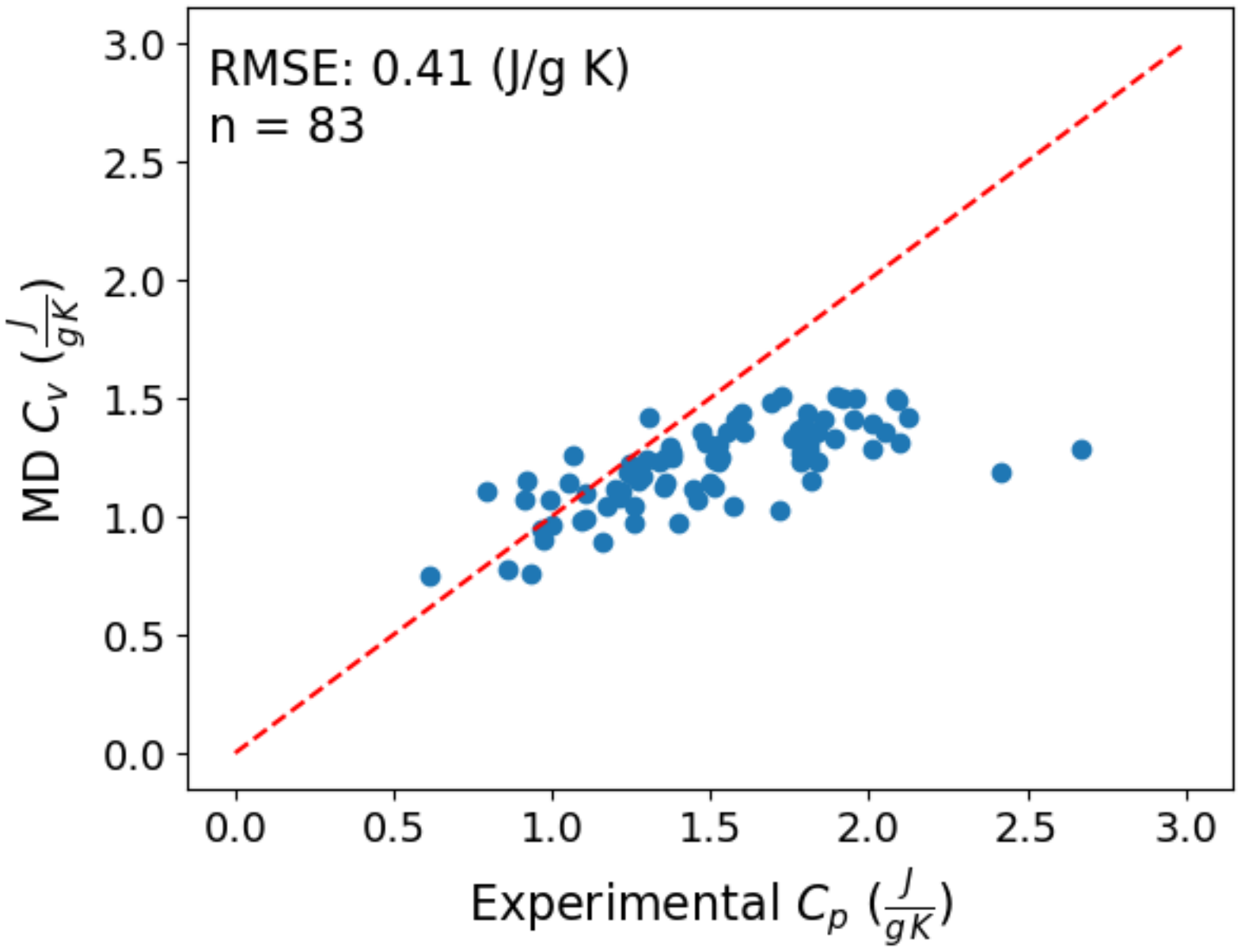

**Figure S6**: Comparison between quantum-corrected MD-calculated $C_v$ (OPoly26) and experimental specific heat capacity values from PolyInfo. $r^2 = 0.50$. The dashed red line indicates y=x.

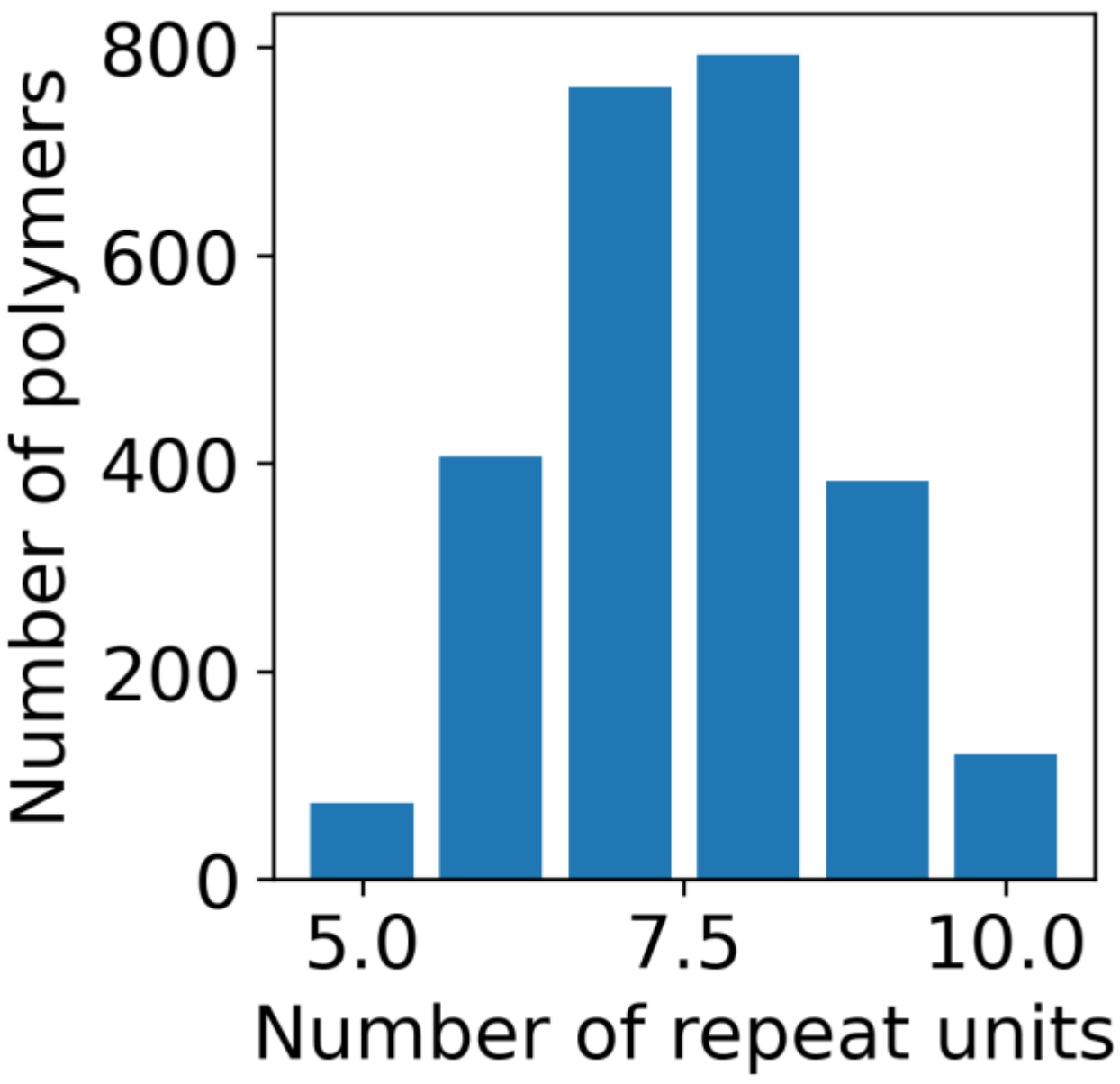

**Figure S7**: Number of data points from MD simulations of refractive index as a function of number of distinct repeat units per polymer.

**Table S3.** SMILES/English ChatGPT prediction results

| Model | Electron Affinity (DFT) RMSE (eV) | $T_g$ RMSE (°C) |
|---|---|---|
| GPT-4o, SMILES | 1.20 | 73.5 |
| GPT-4o, English | 1.28 | 83.1 |
| GPT-o3-mini, SMILES | 0.95 | 70.4 |
| GPT-o3-mini, English | 1.09 | 72.2 |
| PC | **0.29 ± .05** | **34.7 ± 0.6** |
| PPG | **0.29 ± .06** | 36.7 ± 0.8 |

| polyBERT | 0.39 ± .05 | 37.9 ± 1.8 |
|---|---|---|
| RF | 0.43 ± .06 | 39.7 ± 1.5 |

## GPT example prompts

Prompt 1: No naming

What's the density of this polymer? `*CCC(=O)N*`. The PSMILES strings contains asterisks (`*`), these are part of the molecule and must not be interpreted as formatting or markdown.

Explain your reasoning, but do not reference the common or English name of this polymer.

End your response with the predicted value on a separate line, in this format:

`Density: <float> g/mL`.

Prompt 2: Required naming

What's the density of this polymer? `*CCC(=O)N*`. The PSMILES strings contains asterisks (`*`), these are part of the molecule and must not be interpreted as formatting or markdown.

Explain your reasoning, and explicitly give the common or english name for this polymer.

End your response with the predicted value on a separate line, in this format:

`Density: <float> g/mL`.

## Training time details

**Table S4.** Training time comparison between the different models trained on the same 1,800 data point homopolymer density dataset.

| Model | Training time (s) |
|---|---|
| PC | 274 |
| PPG | 362 |
| polyBERT | 37 |
| RF | 48 |

**Table S5**. Training time comparison between the different models trained on the same 37,133 data point alternating copolymer density dataset.

| Model | Training time (s) |
| --- | --- |
| PC | 34,218 |
| PPG | 35,221 |
| polyBERT | 497 |
| RF | 4,554 |

Training-time measurements were conducted on the LLNL Dane high-performance computing system using Intel Sapphire Rapids CPU. Each training run used 1 CPU core without GPU acceleration. We report wall-clock time for model fitting, including feature generation (where applicable) and validation. All models were evaluated using identical data splits and training-set sizes.